\documentclass[runningheads]{llncs}
\usepackage{graphicx}
\usepackage{caption}
\usepackage{subcaption}
\usepackage{listings}
\begin{document}
\title{Code quality assessment using transformers}
%
%\titlerunning{Abbreviated paper title}
% If the paper title is too long for the running head, you can set
% an abbreviated paper title here
%
\author{Mosleh Mahamud\inst{1}\and
Isak Samsten\inst{1}}
\authorrunning{M. Mahamud and I. Samsten}
% First names are abbreviated in the running head.
% If there are more than two authors, 'et al.' is used.
%
\institute{Department of Computer and Systems Sciences, Stockholm University, Stockholm, Sweden}
\maketitle              % typeset the header of the contribution
\begin{abstract}
Automatically evaluate the correctness of programming assignments is rather straightforward using unit and integration tests. However, programming tasks can be solved in multiple ways, many of which, although correct, are inelegant. For instance, excessive branching, poor naming or repetitiveness make the code hard to understand and maintain. These subjective qualities of code are hard to automatically assess using current techniques. In this work we investigate the use of CodeBERT to automatically assign quality score to Java code. We experiment with different models and training paradigms. We explore the accuracy of the models on a novel dataset for code quality assessment. Finally, we assess the quality of the predictions using saliency maps. We find that code quality to some extent is predictable and that transformer based models using task adapted pre-training can solve the task more efficiently than other techniques.   

\end{abstract}
\section{Introduction}
Grading of assignments is one of the prevalent ways in which students obtain feedback on their work\cite{rahman2007review}. Such grading requires massive amounts of repetitive work from examiners to give grades on assignments to students. Programming assignments are a common form of examination in teaching programming at higher education. Despite the fact that the number of students at on-campus courses rarely exceeds 500, with the growing interest and utilization of massive open online courses (MOOC), the number of students can range from a few hundred to many thousands. With large student cohorts, giving timely and useful feedback becomes prohibitively costly. A multitude of automated tools have been introduced to correct and grade programming assignments. These tools are often characterized as either dynamic or static \cite{rahman2007review}. Static systems perform code analysis without running the actual assignments. As such, these techniques can be used to detect syntax errors and a small family of semantic errors. Moreover, some quantitative measures of code quality can also be computed such as the number of lines of code, number of exit-points in a function and so on. Conversely, dynamic systems evaluate the assignments by running the code and through multiple test cases ensure that the functional specification of the assignment is correct. 

There are many ways to solve such assignments, leading to a large number of possible solutions that although correct are not (subjectively) elegant and maintainable. The large variation of programming solutions are almost impossible to assess using traditional unit testing methods \cite{heckman2018developing}. This makes it a laborious process for the course's lecturer to grade students' assignments manually.  Therefore, automated process is of strong desire that can achieve this in scale and within time. This widens the opportunity to apply machine learning and other supervised learning techniques in this scenario.

Ever since the first transformer architecture \cite{vaswani2017attention} was proposed, significant attention has been given to identifying different architectures and pre-trained models. For example, BERT \cite{devlin2018bert} and RoBERTa \cite{liu2019roberta} has been introduced as immensely successful architectures during the last years. In fact, these models have achieved exceptional levels of language representation and have also proven to be effective at many different downstream tasks within Natural Language Processing (NLP). There have also been models that work with different modalities such as video and audio \cite{su2019vl,sun2019utilizing}. Recently there have been many attempts to do the same with Code Language models similar to GPT-3\cite{brown2020language}, CodeGPT\cite{lu2021codexglue}, where these architectures have been used to create products such as Github Copilot and Codex by Open AI. Moreover, novel encoder models such as CodeBERT\cite{feng2020codebert} and Graph-CodeBERT\cite{guo2020graphcodebert} have recently been introduced in attempts to do code to natural language generation tasks, code documentation and natural language code search.

Making effective use of transformers often require extensive pre-training it with additional data. In particular, pre-training on task and domain related data is important since the model can learn from the extra data that is within the subject of the downstream task. In the literature, two main categories of pre-training has been described \cite{gururangan2020don}. Domain Adaptive Pre-training (DAPT) and Task Adaptive Pre-training (TAPT), where the former is about training domain related general data related to the downstream task and the latter is about pre-training using task specific data that is closer to the down stream tasks with unlabeled data.

\subsection{Contributions}
In this paper, we propose using a CodeBERT model for assessing the code quality of Java  methods and apply this to assessing the quality of programming assignments. Moreover, we explore and fine-tune CodeBERT for the novel task of assessing code quality. Moreover, i several experiments we hightlight the importance of pre-traning and the need for transformer based models that consider the complex semantics of programming languages. We make the following contributions:

\begin{itemize}

\item We explore the novel task of predicting code quality using transformers and also contribute a novel dataset for the task.

\item We also study effects of proximal pre training to downstream task on isolated transformer model.

\item We study the impact of task adapted and domain adaptive pre-training and show that transformer based models significantly outperform baseline approaches.

%\item We suggest and explore ways of providing explanations for the code quality predictions.
\end{itemize}

\section{Experimental evaluation}
We conduct our experiments using a novel dataset, consisting of a Java programming exercise solved by students at Stockholm university during 2019. The dataset consist of programming solutions for a JavaFX task completed by 250 students, which corresponds to more than 25000 java methods. For each method declaration, domain experts assigns a score based on the methods intrinsic quality. This score is subsequently used to represent the quality of the method. While several quantitative measures of code quality (e.g., number of lines) has been proposed in the literature, the quality score assigned by the domain expert represents the subjective quality of the code, .e.g, elegance, readability and succinctness. In the experiments, we partition the score into two labels indicating \emph{good quality} and \emph{bad quality} code. The partitioning results in a class distribution of 70\% good quality methods and 30\% bad quality methods. In total, we labeled 2500 methods which where subsequently divided into three parts for training, validation and testing. The predetermined split is consistent with different models for training and testing.

\subsection{Experimental setup}
In the experiments, we include two different classifiers: a baseline using Random Forest\cite{breiman2001random} and CodeBERT which is configured with either, no extra pre-training, domain adapted pre-training, task-adapted pre-training or with task-only-training. For each instance of CodeBERT, we add a softmax layer as the output layer and set the hyper-parameters to: learning rate to $2^{-5}$, batch size to $16$ and epochs to $3$, with a weight decay of $0.01$. Finally, we use the Adam optimizer to train the model.

\subsubsection{Baseline.} As a baseline we use a random forest model (with 100 trees) with each code sample represented as a vector of term-frequency inverse document-frequency for each token in the corpus. We denote the baseline as \texttt{TFIDF-RF}.

\subsubsection{CodeBERT.} CodeBERT is a multi-layer bidirectional Transformer, similar to Bert \cite{devlin2018bert} and RoBERTa \cite{liu2019roberta}, which has been pre-trained for multiple code related NLP tasks. As stated above, we augment the architecture with a linear softmax-layer on top of the pooled output. We will not review the ubiquitous Transformer architecture in detail but refer the interested reader to \cite{devlin2018bert}.  In addition to the default CodeBERT (\texttt{CodeBERT-Base}), we pre-trained CodeBERT using masked language modelling, similar to Roberta\cite{liu2019roberta}, either domain adapted (\texttt{DAPT-CodeBERT}) or task adapted (\texttt{TAPT-CodeBERT}). The domain adapted model is a CodeBert that is further pre-trained with an additional 200,000 Java methods downloaded from GitHub repositories tagged with JavaFX. Similarly, task adapted model is another CodeBERT that is further  pre-trained with 10,000 Java methods from previous years JavaFX assignment. Because the task-adapted dataset only contains 10,000 samples, we adapted the hyper-parameters suggested by \cite{gururangan2020don} during per-training. However, contrary to \cite{gururangan2020don}, we do not augment the pre-training data but instead train both models using the hyper-parameters in Table~\ref{tab:hyp}. This is done to observe the performance difference of transformer models with a variety of pretraining and how much a pretraining data's proximity to downstream tasks affects the performance of models. As per our knowledge, no previous research shows this in code language models.   Finally, we pre-train a model only using the task-adapted data, i.e., without any additional pre-training. We call this model \texttt{FxBERT}. This FxBERT model is a 6 layer encoder from transformers with the same configurations as CodeBERT. FxBERT has 6 layers instead of 12 compared to CodeBERT to make it space effitient and possible inference speed up. The hyperparameters were the same as TAPT-CodeBERT for pre-training using masked language modeling. This is done to see if pre training with other code languages on a typical CodeBERT is needed to solve a specific downstream task. This isolated pretraining is unorthodox in the language model space and is usually discouraged (cite) but this paper challenges this idea as we don't know the limitations of code language models and how they react to pretraining with small datasets.

%\begin{table}
%\caption{The different parameters used for training, where $\epsilon$ values refer to the Adam optimizer which was used during training.\label{tab:hyp}}
%\centering
%\begin{tabular}{ l | r  r }
%    \hline
%    \textbf{  } &     \texttt{DAPT-CodeBERT} &   \texttt{TAPT-CodeBERT}  \\
%    \hline
%         Epochs &        3 &      100 \\ 
%    Batch Size &        8 &        8 \\
%    Learning Rate &     $1^{-5}$ &     $1^{-4}$ \\
%        $\epsilon$ & $1^{-8}$ &   $1^{-8}$ \\
%\hline  
%\end{tabular}
%\end{table}

To evaluate the performance of the different architectures we report the accuracy (i.e., the number of correctly classified samples). Since the class-distribution is skewed, we also report the area under the ROC-curve, which measure the models ability to rank a true positive sample ahead of a negative sample. Finally, we also report the area under precision-recall-curve which measure the precision at different levels of recall as well as the average F1-score, i.e., the harmonic mean between precision and recall. Finally, all the models were trained on a V100-SMX2 with 16 GB of memory on google colab.

\section{Empirical Investigation}
In this section, we compare the different CodeBERT architectures both internally but also compared to the baseline approach using random forest. In Table~\ref{tbl:res}, we can see that due to the class imbalance, all methods perform similarly when considering accuracy. The only difference we can note is that the task-adapted CodeBERT model perform slightly better than the alternatives, with 86\% accuracy compared to 85\% for the baseline and 83\% for the default CodeBERT architecture.  Similarly, considering FxBERT is only pre-trained with task adaptive data it still outperformed CodeBERT-base by 2\% on the accuracy, whilst still being 35\% smaller than other Code-BERT models.

\begin{table}[]
\caption{The full classification result for all methods\label{tbl:res}}
\centering
\begin{tabular}{l | rrrrrr}
\hline
                       & \textbf{Precision} & \textbf{Recall} & \textbf{F1}   & \textbf{Accuracy} & \textbf{AUCROC} & \textbf{AUPRC} \\\hline
\texttt{TFIDF-RF}      & 0,77               & 0,66            & 0,69          & 0,85              & 0,707           & 0,898          \\
\texttt{CodeBERT-Base} & 0,72               & 0,66            & 0,68          & 0,83              & 0,728          & 0,915          \\
\texttt{DAPT-CodeBERT} & 0,76               & 0,67            & 0,7           & 0,84              & 0,724          & 0,910          \\
\texttt{TAPT-CodeBERT} & \textbf{0,81}      & \textbf{0,68}   & \textbf{0,72} & \textbf{0,86}     & \textbf{0,741} & \textbf{0,919} \\
\texttt{FxBERT}        & 0,76               & \textbf{0,68}   & 0,71          & 0,85              & 0,704          & 0,905          \\\hline
\end{tabular}
\end{table}

Interestingly, all CodeBERT based models have higher AUROC and AUPRC than the baseline method, which indicates that the probabilities assigned by these methods are more well-calibrated than those of the baseline. Similarly, if we inspect the (average) precision we can see that the task-adapted CodeBERT model significantly outperforms the other models. The same conclusion also holds for (average) recall and the (average) harmonic mean between precision and recall. 
However, CodeBert-Base performed comparatively worse than BoW baseline and other models questioning if Codebert is needed to be pre trained with other code languages, especially when compared to FxBERT which is a smaller architecture and pre trained with minute amounts of data. Moreover, it is hard to understand if a large parameter model is needed to make a decent code language model as code languages have fewer nuances than natural language.

\begin{figure}
\begin{subfigure}[b]{0.49\textwidth}
    \includegraphics[width=\textwidth]{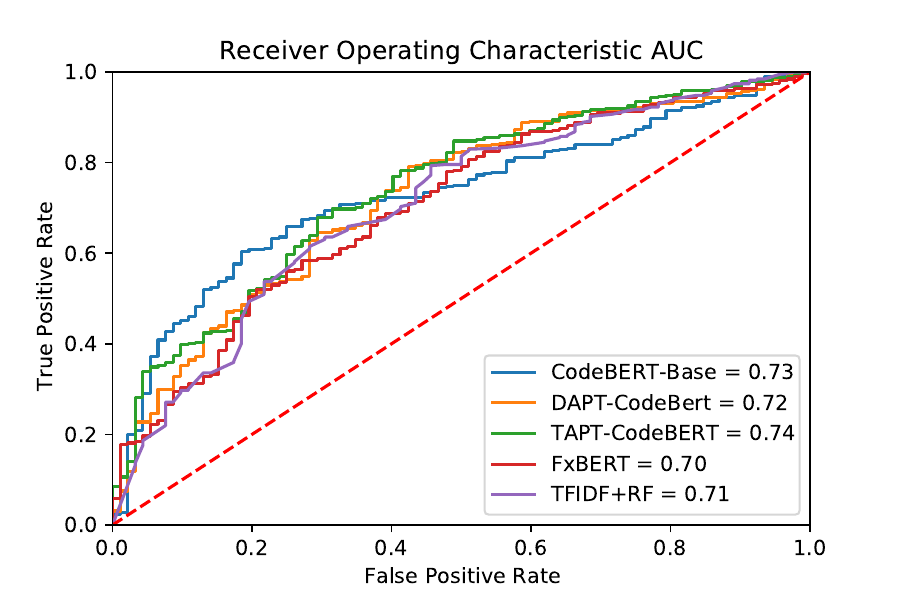}
    \caption{The receiver-operator curve for all CodeBERT architectures and the baseline.}
    \label{fig:f1}
  \end{subfigure}
  \hfill
  \begin{subfigure}[b]{0.49\textwidth}
    \includegraphics[width=\textwidth]{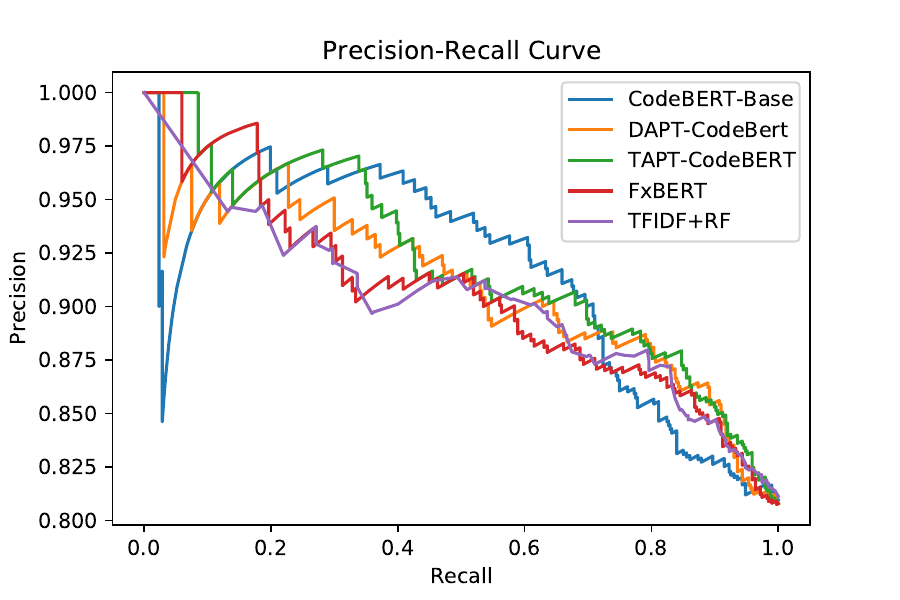}
    \caption{The precision-recall curve for all CodeBERT architectures and the baseline.}
    \label{fig:f2}
  \end{subfigure}
\end{figure}

Finally, in Figure~\ref{fig:f1} we see the ROC curve for all methods and can clearly see that for low false-positive rate the default CodeBERT model has the highest true positive rate, but as the false-positive rate increases, the importance of pre-training is highlighted. In particular, we can see that the domain- and task-adapted models perform better than the default CodeBERT model. Similarly, in Figure~\ref{fig:f2} we see that for low recall, the precision of all CodeBert models is low, but as recall increases so does the rate of decline of precision.

\subsection{Examples and interpretability}
\begin{lstlisting}[language=Java, basicstyle=\ttfamily\footnotesize, frame=lines, caption={Example of a method predicted as bad code quality.}]
public void clearCenter() { 
    for(Entry<Position, Triangle> a : pt.entrySet()) { 
        Triangle t = a.getValue(); 
        center.getChildren().remove(t);
    } 
    cl.clear(); namedList.clear(); pt.clear()
}
\end{lstlisting}

To interpret the predictions by the best performing model, i.e., TAPT-CodeBERT, we use Shap\cite{lundberg2017unified} and compute the attribution of each feature towards either of the predicted labels. In Listing 1.1, we show an example of a method that was predicted as bad quality by TAPT-CodeBERT, and the explanation is shown in Table~\ref{tbl:attr}, which shows a sample of the highest and lowest attribution scores produced by Shap. By inspecting the attribution scores, we can see that the token n-gram with highest attribution is the use of the method \texttt{entrySet()} which has an attribution of $0.136$ and the most important token n-gram which is negatively affecting the prediction is the \texttt{Triangle t = a .}. The explanation does indeed highlight one of the main drawbacks of the implementation, namely the use of \texttt{entrySet{}} while iterating over all key-value pairs while only utilizing the value a more elegant solution would be to iterate over the values, i.e., use the method \texttt{values()}. Moreover, two of the calls to \texttt{clear()} are also highlighted as important for the prediction. The clearing of multiple lists seems like an anti-pattern and they should probably be encapsulated in a single aptly named method or class.

\begin{table}[]
\centering
\caption{Highest and lowest attributions for predicting the label \emph{bad code} for the method in Listing~1.1\label{tbl:attr}}
\begin{tabular}{l|r}
\hline 
\textbf{Token n-gram}                                     & \textbf{Attribution}\\ \hline
\texttt{Entry\textless{}Position}                         & 0.023             \\
\texttt{Triangle }                                        & -0.008            \\
\texttt{entrySet()) \{}                                   & 0.136             \\
\texttt{Triangle t = a . }                                & -0.058            \\
\texttt{getValue(); center.getChildren().remove}.        & 0.051             \\
\texttt{\} cl.clear()}                                         & 0.015             \\
\texttt{; namedList.clear()}                          & 0.03             
\end{tabular}

\end{table}

\section{Conclusions}
In this paper, we explore the novel task of assessing the quality of code and apply this task to the grading of programming assignments. In several experiments, we evaluate the performance of five different training schemes for CodeBERT, which is a state-of-the-art model for code prediction, and show that the performance significantly outperforms that of a simple baseline. Furthermore, we explore different pre-training approaches and show that using task-adapted pre-training can improve performance on the task of predicting code quality. For future work, it would be important to expand the scope of the dataset to not use a binary output but instead model the quality as a score. Moreover, while pre-training indeed improves the importance, for future work it is important to try and understand how the model architecture can be improved to facilitate better predictions. Finally, we would like to explore how the scores for each Java method in a program can be combined to assess the quality of a complete program.

\subsubsection{Code \& Dataset availability}
All code and dataset will be available at the supporting GitHub repository \url{https://github.com/mosh98/AutoGrade}.

\bibliographystyle{splncs04}
\bibliography{references}
\end{document}